%% file: main.tex
\documentclass[10pt,twocolumn,letterpaper]{article}

\usepackage{cvpr}              

\input{preamble}

\definecolor{cvprblue}{rgb}{0.21,0.49,0.74}
\usepackage[pagebackref,breaklinks,colorlinks,citecolor=cvprblue]{hyperref}
\def\paperID{*****} 
\def\confName{CVPR}
\def\confYear{2024}

\title{TrafficBots V1.5: Traffic Simulation via Conditional VAEs \\ and Transformers with Relative Pose Encoding
}

\author{Zhejun Zhang \\
ETH Z\"urich\\
{\tt\small zhejun.zhang@vision.ee.ethz.ch}
\and
Christos Sakaridis\\
ETH Z\"urich\\
{\tt\small csakarid@vision.ee.ethz.ch}
\and
Luc Van Gool\\
ETH Z\"urich, INSAIT Sofia\\
{\tt\small vangool@vision.ee.ethz.ch}
}

\begin{document}
\maketitle
\input{sec/0_abstract}    
\input{sec/1_intro}
\input{sec/2_method}
\input{sec/3_results}
\input{sec/4_conclusion}
\clearpage
{
    \small
    \bibliographystyle{ieeenat_fullname}
    \bibliography{main}
}


\end{document}

%% file: preamble.tex
\usepackage[dvipsnames]{xcolor}


%% file: sec/0_abstract.tex
\begin{abstract}
In this technical report we present TrafficBots V1.5, a baseline method for the closed-loop simulation of traffic agents.
TrafficBots V1.5 achieves baseline-level performance and a 3rd place ranking in the Waymo Open Sim Agents Challenge (WOSAC) 2024.
It is a simple baseline that combines TrafficBots, a CVAE-based multi-agent policy conditioned on each agent's individual destination and personality, and HPTR, the heterogeneous polyline transformer with relative pose encoding.
To improve the performance on the WOSAC leaderboard, we apply scheduled teacher-forcing at the training time and we filter the sampled scenarios at the inference time.
The code is available at: \url{https://github.com/zhejz/TrafficBotsV1.5}

\end{abstract}

%% file: sec/1_intro.tex
\section{Introduction}

The problem of closed-loop multi-agent traffic simulation can be addressed by learning a policy for each traffic participants.
Specifically, at each time step, the policy predicts the next action of each agent, given the historical observations form previous time steps, including the map, traffic lights and agent trajectories.
Based on TrafficBots~\cite{zhang2023trafficbots}, the TrafficBots V1.5 policy is shared by all agents.
Different behaviors are generated by conditioning the policy on the individual destination and personality of each agent.
In contrast to the SceneTransformer~\cite{ngiam2021Scene} network architecture and input representation, which are not rotation and translation invariant, TrafficBots V1.5 uses the pairwise-relative representation and the HPTR~\cite{zhang2024real} architecture.
This greatly improves the accuracy of TrafficBots without sacrificing its efficiency and scalability.
Moreover, instead using a recurrent neural network (RNN) to encode the temporal axis, TrafficBots V1.5 uses stacked historical observation as input, such that its architecture is solely based on Transformers~\cite{vaswani2017attention}.

\subsection{TrafficBots}

TrafficBots~\cite{zhang2023trafficbots} is a multi-agent policy built upon motion prediction and end-to-end (E2E) driving.
Compared to previous data-driven traffic simulators~\cite{suo2021trafficsim}, TrafficBots demonstrates superior configurability and scalability.
To generate configurable behaviors, for each agent TrafficBots introduces a destination as navigational information, and a time-invariant latent personality that specifies the behavioral style.
Unlike the goal which depends on the prediction horizon and hence leads to causal confusions, the destination approximates the output of a navigator which is available in the problem formulation of E2E driving~\cite{zhang2021end}.
Importantly, the destination indicates where the agent wants to reach eventually, i.e., not necessarily at a specific future time step.
In order to capture the diverse behaviors from human demonstrations, the personality is learned using the conditional variational autoencoder (CVAE)~\cite{sohn2015learning} following prior works on multi-modal motion prediction~\cite{casas2020implicit}.
To ensure the scalability, TrafficBots uses the scene-centric representation~\cite{ngiam2021Scene} and presents a new scheme of positional encoding for angles, allowing all agents to share the same vectorized context and the use of an architecture based on Transformers.
However, due to the lack of rotation and translation invariance induced by the scene-centric representation, TrafficBots does not achieve superior performance compared to methods using the agent-centric representation.

\subsection{HPTR}

Depending on how the coordinate system of the vectorized representation is selected, motion prediction methods fall into three categories: agent-centric~\cite{nayakanti2022wayformer}, scene-centric~\cite{ngiam2021Scene} and pairwise-relative~\cite{cui2022gorela}.
While agent-centric methods achieve top accuracy but lack scalability, and scene-centric methods show superior scalability but suffer from poor accuracy, the pairwise-relative methods get the best of both worlds.
However, previous pairwise-relative methods are mostly using graph neural networks~\cite{cui2022gorela}, which are often less efficiently implemented on the graphics processing unit (GPU) compared to Transformers with dot-product attention.
To address this problem, a novel attention module called \textbf{K}-nearest \textbf{N}eighbor \textbf{A}ttention with \textbf{R}elative \textbf{P}ose \textbf{E}ncoding (\textsc{Knarpe}) is introduced in~\cite{zhang2024real}, which allows the pairwise-relative representation to be used by Transformers.
\textsc{Knarpe} projects the relative pose encoding (RPE) and adds them to the keys and values to obtain $\mathbf{z}_i$, the output of letting token $i$ attend to its $K$ nearest neighbors $\kappa_i^K$,
\begin{align}
    \mathbf{z}_i &= \textsc{Knarpe}\left(\mathbf{u}_i,\mathbf{u}_j, \mathbf{r}_{ij} \mid j \in \kappa_i^K \right) \\
    &=\sum_{j\in \kappa_i^K}\alpha_{ij} \left(\mathbf{u}_j\mathbf{W}^v+\mathbf{b}^v{+\text{RPE}(\mathbf{r}_{ij})\mathbf{\hat W}^v+\mathbf{\hat b}^v} \right)  \\
    \alpha_{ij} &=\frac{\exp(e_{ij})}{\sum_{k\in \kappa_i^K}\exp(e_{ik})} \\
    e_{ij} &= \frac{(\mathbf{u}_i\mathbf{W}^q+\mathbf{b}^q)(\mathbf{u}_j\mathbf{W}^k+\mathbf{b}^k{+\text{RPE}(\mathbf{r}_{ij})\mathbf{\hat W}^k+\mathbf{\hat b}^k})}{\sqrt{D}}
\end{align}
where $\mathbf{u}_i, \mathbf{u}_j$ are the local attributes of token $i$ and $j$ represented in their local coordinates, $\mathbf{r}_{ij}$ is the relative pose between token $i$ and $j$, $\alpha_{ij}$ are the attention weights, $e_{ij}$ are the logits, $W^{\{q,k,v\}}, b^{\{q,k,v\}}$ are the learnable projection matrices and biases for query, key and value, and $\hat W^{\{k,v\}}, \hat b^{\{k,v\}}$ are the learnable projection matrices and biases for RPE.
The RPE is defined as follows:
\begin{align}
    \text{RPE}(\mathbf{r}_{ij}) & = \text{concat}(\text{PE}(x_{ij}) \text{PE}(y_{ij}), \text{AE}(\theta_{ij}))  \\
    \text{PE}_{2i}(x) & = \sin(x\cdot\omega^{\frac{2i}{D}}) \\
    \text{PE}_{2i+1}(x) &= \cos(x\cdot\omega^{\frac{2i}{D}}) \\
    \text{AE}_{2i}(\theta) & = \sin\left(\theta\cdot (i+1)\right)\\
    \text{AE}_{2i+1}(\theta) &= \cos\left(\theta\cdot (i+1)\right)\\
    i&\in\{0,\dots,D/2-1\},
\end{align}
where $(x_{ij},y_{ij},\theta_{ij})$ are the 2D location and yaw heading of token $j$ represented in the coordinate of token $i$, $\omega$ is the base frequency, and $D$ is the embedding dimension.

Based on \textsc{Knarpe}, a pure Transformer-based framework called \textbf{H}eterogeneous \textbf{P}olyline \textbf{T}ransformer with \textbf{R}elative pose encoding (HPTR)~\cite{zhang2024real} is presented, which uses a hierarchical architecture to enable asynchronous token update and avoid redundant computations.
By sharing contexts among traffic agents and reusing the unchanged contexts in driving scenarios, HPTR is as efficient as scene-centric methods, while performing on par with state-of-the-art agent-centric methods on marginal motion prediction tasks.
Since our method TrafficBots V1.5 is entirely based on \textsc{Knarpe} and HPTR, we refer the readers to the HPTR~\cite{zhang2024real} paper for more details about the mathematical formulation and the network architecture.

%% file: sec/2_method.tex
\section{Method}

\begin{figure*}[t]
    \centering
    \includegraphics[width=0.95\textwidth]{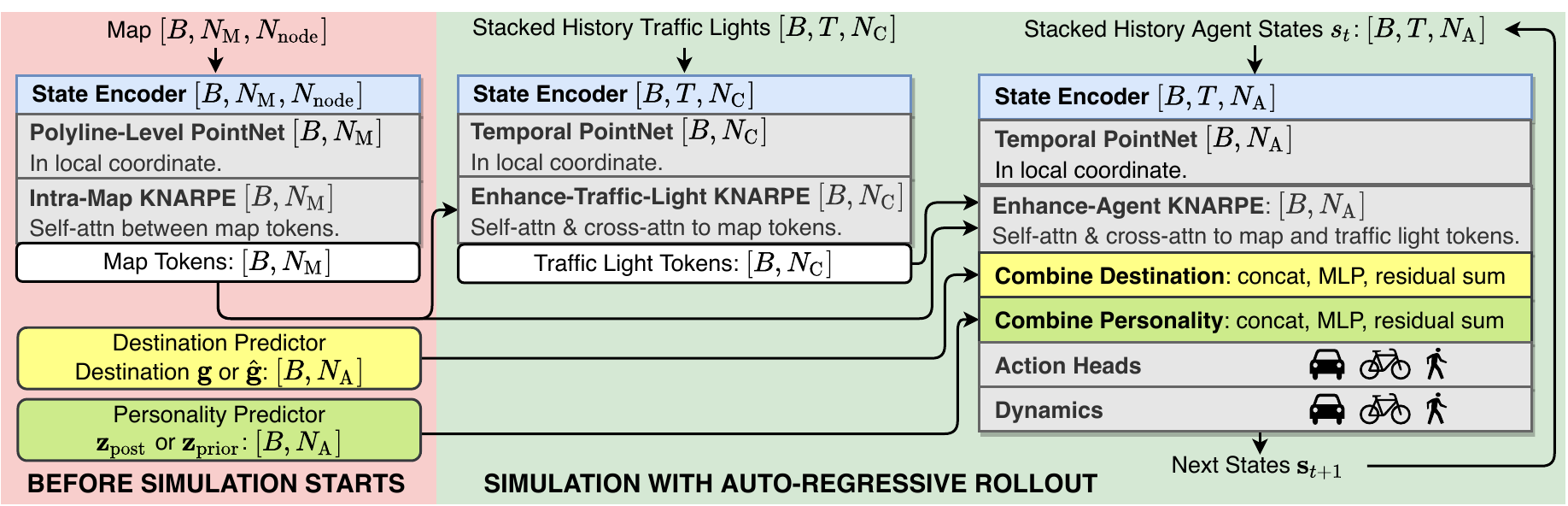}
    \vspace{-0.5ex}
    \caption{\textbf{Network architecture of TrafficBots V1.5}. In the brackets are the tensor shapes, where the hidden dimensions are omitted for conciseness. $B$ is the batch size, which is also the number of episodes. $N_\text{M}, N_\text{C}, N_\text{A}$ are, respectively, the number of map polylines, traffic light polylines and agent trajectories. $N_\text{node}$ is the number of segments in each polyline. $T$ is the length of the stacked historical observations. The destination predictor and personality predictor are not visualized. They have a similar structure to the policy network.
    }
    \label{fig:network}
    \vspace{-0.5ex}
\end{figure*}

TrafficBots V1.5 updates TrafficBots~\cite{zhang2023trafficbots} with the HPTR~\cite{zhang2024real} input representation and network architecture.
This section describes the changes we have made while combining TrafficBots and HPTR.

\subsection{Architecture}

The network architecture of TrafficBots V1.5 is illustrated in Fig.~\ref{fig:network}.
We make minimal changes while applying HPTR to TrafficBots.
We remove the temporal RNN from TrafficBots, and follow HPTR to use stacked historical observations as input and PointNet~\cite{qi2017pointnet,gao2020vectornet} to aggregate the temporal axis.
The policy network, the personality predictor, and the destination predictor of TrafficBots are now all based on the pairwise-relative representation and \textsc{Knarpe} attention module.
We keep the intra-map, enhance-traffic-light and enhance-agent Transformers of HPTR.
Following TrafficBots, multi-modal outputs are generated by conditioning the policy on each agent's individual destination and personality.
Therefore, the anchors and the anchor-to-all Transformer of HPTR are discarded.
Instead of a learnable prior personality of TrafficBots, we use a standard Gaussian for the prior personality.
We also tried to add a traffic light state predictor, but its accuracy was not good enough to improve the overall simulation performance on the leaderboard.

\subsection{Training}

We found two techniques that improve the performance of TrafficBots V1.5.
Firstly, we adopt a larger free nats~\cite{hafner2019Learning} equals to $1.0$ for the KL-divergence between the posterior and the prior personality.
This allows the posterior personality to encode more information.
Secondly, we use scheduled sampling~\cite{bengio2015scheduled} and apply teacher-forcing to $30\%$ agents at the beginning of the training, i.e., these agents will be trained via open-loop behavior cloning.
The percentage of teacher-forcing decreases linearly to $0$ during the training.
Due to limited computational resources, we train our models for 5 days on 4 NVIDIA RTX 4090 GPUs.
Training for longer time or using more GPUs should improve the performance further.
We use a total batch size of $8$ and we batch over scenarios.
Each scenario contains $91$ time steps and a maximum of $64$ agents.
Following TrafficBots, the training uses back-propagation through time and the training loss includes the following terms:
\begin{enumerate}
  \item Reconstruction loss that trains the model to reconstruct the ground-truth (GT) states using the posterior personality and the GT destination. It is a weighted sum of:
  \begin{itemize}
  \item A smoothed L1 loss between the predicted and the GT $(x,y)$ positions.
  \item A cosine distance between the predicted and the GT yaw heading.
  \item A smoothed L1 loss between the predicted and the GT velocities.
  \end{itemize}
  \item The KL divergence between the posterior and the prior personality. We use free nats~\cite{hafner2019Learning} to clip the KL divergence, i.e., if $KL(\mathbf{z}_\text{post}, \mathbf{z}_\text{prior})$ is smaller than the free nats, then the KL loss is not applied.
  \item The cross entropy loss for destination classification. Since the GT destination is a single class, this loss boils down to a maximum likelihood loss, i.e., the destination distribution is trained to maximize the log-likelihood of the polyline index of the GT destination.
\end{enumerate}
During training, we auto-regressively rollout the policy using the GT destination and the posterior personality.
Before the auto-regressively rollout, we use the reparameterization trick to sample the posterior personality, such that it can be trained to reconstruct the GT trajectories.
For $10\%$ of the episodes we rollout with the prior personality instead of the posterior personality.
During the auto-regressively rollout, we take the mean of the actions such that the gradient can be back-propagated through the differentiable vehicle dynamics.
We do not apply a loss that encourages collision avoidance~\cite{cui2022gorela, lu2023imitation, zhang2023learning} because this will bias the model, even though it could improve the performance on the leaderboard.

\begin{table*}
\setlength{\tabcolsep}{10pt}
\centering
\caption{Results on the WOSAC leaderboard 2024~\cite{wosac2024leaderboard}.}
\label{table:wosac}
\begin{tabular}{lcccccc} 
\toprule
Method name
& \begin{tabular}{@{}c@{}} Realism meta \\ metric $\uparrow$ \end{tabular} 
& \begin{tabular}{@{}c@{}} Kinematic \\ metrics $\uparrow$ \end{tabular} 
& \begin{tabular}{@{}c@{}} Interactive \\ metrics $\uparrow$ \end{tabular} 
& \begin{tabular}{@{}c@{}} Map-based \\ metrics $\uparrow$ \end{tabular} 
& \begin{tabular}{@{}c@{}} minADE \\ $\downarrow$ \end{tabular}  \\
\cmidrule(lr){1-1}\cmidrule(lr){2-6}
 SMART-large~\cite{wu2024smart} &
$0.7564$ & $0.4769$ & $0.7986$ & $0.8618$ & $1.5501$ \\
BehaviorGPT~\cite{zhou2024behaviorgpt} &
$0.7473$ & $0.4333$ & $0.7997$ & $0.8593$ & $1.4147$ \\
GUMP &
$0.7431$ & $0.4780$ & $0.7887$ & $0.8359$ & $1.6041$ \\
model\_predictive\_submission &
$0.7417$ & $0.4182$ & $0.7942$ & $0.8591$ & $1.4842$ \\
MVTE~\cite{wang2023multiverse} &
$0.7302$ & $0.4503$ & $0.7706$ & $0.8381$ & $1.6770$ \\
VBD &
$0.7200$ & $0.4169$ & $0.7819$ & $0.8137$ & $1.4743$ \\
TrafficBots V1.5 (Ours) &
$0.6988$ & $0.4304$ & $0.7114$ & $0.8360$ & $1.8825$ \\
cogniBot v1.5 &
$0.6288$ & $0.3293$ & $0.7129$ & $0.6918$ & N/A \\
linear\_extrapolation\_baseline &
$0.3985$ & $0.2253$ & $0.4327$ & $0.4533$ & $7.5148$ \\
\bottomrule
\end{tabular}
\end{table*}

\subsection{Inference}

Instead of bias the model directly, we apply a milder approach to bias the model's outputs towards safer behavior and hence improve the collision-based metrics on the WOSAC leaderboard.
Specifically, we sample $128$ scenarios at the inference time and select $32$ scenarios that contain the least collision events.
To sample a scenario, we first sample the personality and destination for each agent, after that we start the auto-regressive rollout.
We use the mean of the predicted action distribution, hence the rollout is completely deterministic given the sampled personality and destination.
Except the episode filtering, we do not apply any other post-processing or model ensembling techniques.
At the inference time, HPTR allows different types of tokens to be updated asynchronously at different frequency.
Therefore, an accurate analysis of FLOPS turns out to be difficult.
For simplicity, we only profile the map encoder, which is the computationally heaviest module in our model.
For a single episode, the map encoder uses 20 GFLOPS.
Since HPTR allows the map features to be cached and reused during the rollout, the map is encoded only once before the simulation starts.
Other modules, e.g. the personality encoder, the traffic light encoder and the policy, are computationally much lighter than the map encoder.
Based on the FLOPS of the map encoder, we estimate that each iteration of the auto-regressive policy rollout should require approximately an order of magnitude fewer FLOPS, i.e., around 2 GFLOPS.

\subsection{Implementation details}
Overall, we use a hidden dimension of $128$.
We use Transformer with pre-layer normalization, and the attention module has $4$ heads and a feed-forward dimension of $512$.
The sampling rate is fixed to $10$ FPS at both the training and inference time.
At each time step, our method predicts the actions one step ahead, based on the previous observations.
We do not differentiate between the ego-agent, i.e., the autonomous vehicle agent, and other agents.
Similar to TrafficBots, TrafficBots V1.5 allows agents that are not controlled by itself.
Since TrafficBots V1.5 predicts the action for all observed agents, the actions will still be predicted for those agents, but then these actions will be discarded; the behavior of those agents will be overridden by some other control modules, e.g. an AV planner software or log-reply.
Each episode contains at most $1024$ map tokens, $128$ traffic light tokens and $64$ agent tokens, where invalid tokens are masked.
Each map token corresponds to a polyline consisting of up to 20 segments, each 1 meter in length.
Each traffic light token corresponds to the polyline it associated to and the traffic light state.
Each agent token corresponds to an agent trajectory.
We use a sliding window approach and stack the historical observations from the last $11$ steps.
For the Transformers with K-nearest-neighbor attention, each map token attends to the $32$ nearest map tokens.
Each traffic light token attends to $24$ nearest map tokens and $24$ nearest traffic light tokens.
Each agent token attends to $64$ map tokens, $25$ traffic light tokens and $25$ agent tokens in its proximity.
More details about the network architecture can be found in the open-source repository of TrafficBots V1.5, as well as in the TrafficBots~\cite{zhang2023trafficbots} and HPTR~\cite{zhang2024real} papers.
Our model does not take the $z$ axis, i.e., the altitude dimension, into account.
During inference, we assume that the $z$ dimension of agent trajectories remains constant and is equal to its last observed value.

%% file: sec/3_results.tex
\section{Results}

We have not run any ablation studies for our method.
However, we would like to point out some promising directions for ablations, including the discrete action space, weights of different losses, and the collision avoidance loss.
We have done a manual and rough parameter tuning for our model.
The performance of our method on the WOSAC leaderboard~\cite{wosac2024leaderboard} is shown in Table~\ref{table:wosac}.
More details about the challenge and the metrics can be found in the WOSAC paper~\cite{montali2024waymo}.
Our method achieves baseline-level performance in terms of the realism meta metric, which is a weighted sum of other metrics except the minADE.
We apply a unicycle model for the dynamics of all types of agents and select the parameters heuristically.
This allows our simulation to generate smooth trajectories, but it also affects the kinematic metrics negatively.
Our TrafficBots V1.5 is outperformed by other methods in terms of interactive metrics which involve collision avoidance.
We believe adding a loss that encourages collision avoidance would help, but it is controversial if a collision-free simulation would be useful for the development of autonomous driving algorithms.
In terms of map-based metrics that consider off-road driving, our model performs comparably to other methods
The minADE of TrafficBots V1.5 is significantly larger than other methods on the leaderboard, which is a known problem inherited from TrafficBots.
Overall, from Table~\ref{table:wosac} we observe that GPT-based~\cite{radford2019language} architectures that rely on tokenization and next-token prediction, such as SMART and BehaviorGPT, achieve top performance.
Interestingly, none of these GPT-based architectures uses goal or personality conditioning, but they are still able to generate multi-modal outputs.
It seems that the multi-modality in traffic simulation can be addressed using tokenization and the cross-entropy loss.
This indicates that the poor performance of TrafficBots might be caused by the CVAE and regression losses on continuous states and actions.
Another interesting thing is that the GPT-based methods achieve the best performance without considering the traffic lights.
This indicates that the dataset might be imbalanced and the evaluation metrics might be flawed.

%% file: sec/4_conclusion.tex
\section{Conclusion}

In this technical report we present TrafficBots V1.5, which is a baseline method that combines TrafficBots, a prior work on the closed-loop traffic simulation using CVAE, and HPTR, a prior work on the Transformer-based motion prediction using the pairwise-relative representation.
Our method is the only CVAE-based method on the WOSAC leaderboard 2024.
The performance of our method is slightly worse than the GPT-based methods.
However, as a baseline method that involves minor novelty, it achieves the performance we expected, and there are many possibilities to improve this simple baseline.